\documentclass[letterpaper]{article} 
\usepackage{aaai25}  
\usepackage{times}  
\usepackage{helvet}  
\usepackage{courier}  
\usepackage[hyphens]{url}  
\usepackage{graphicx} 
\usepackage{natbib}  
\usepackage{caption} 
\usepackage{algorithm}
\usepackage{algorithmic}
\usepackage{bm}
\usepackage{amsmath}
\usepackage{graphicx}
\usepackage{booktabs}
\usepackage{graphicx}
\usepackage{tabularx}
\usepackage{array}
\usepackage{xcolor}
\usepackage{multirow}
\usepackage{colortbl}
\usepackage{xcolor}
\usepackage{amssymb}
\newcommand\customparagraph[1]{\vspace{0.4em}\noindent\textbf{#1}}

\usepackage{newfloat}
\usepackage{listings}
\DeclareCaptionStyle{ruled}{labelfont=normalfont,labelsep=colon,strut=off} 
\floatstyle{ruled}
\newfloat{listing}{tb}{lst}{}
\floatname{listing}{Listing}
\newcommand{\ie}{i.e.}
\newcommand{\eg}{e.g.}

\title{Adversarial Attacks on Event-Based Pedestrian Detectors: A Physical Approach}
\author{
    Guixu Lin\textsuperscript{\rm 1, \rm 2},
    Muyao Niu\textsuperscript{\rm 1},
    Qingtian Zhu\textsuperscript{\rm 1},
    Zhengwei Yin\textsuperscript{\rm 1},\\
    Zhuoxiao Li\textsuperscript{\rm 1},
    Shengfeng He\textsuperscript{\rm 2},
    Yinqiang Zheng\textsuperscript{\rm 1}\thanks{Corresponding author.}
}
\affiliations{
    \textsuperscript{\rm 1}The University of Tokyo, Japan\\
    \textsuperscript{\rm 2}Singapore Management University, Singapore\\
    
}

\usepackage{bibentry}

\begin{document}

\maketitle

\begin{abstract}
Event cameras, known for their low latency and high dynamic range, show great potential in pedestrian detection applications. However, while recent research has primarily focused on improving detection accuracy, the robustness of event-based visual models against physical adversarial attacks has received limited attention. For example, adversarial physical objects, such as specific clothing patterns or accessories, can exploit inherent vulnerabilities in these systems, leading to misdetections or misclassifications.
This study is the first to explore physical adversarial attacks on event-driven pedestrian detectors, specifically investigating whether certain clothing patterns worn by pedestrians can cause these detectors to fail, effectively rendering them unable to detect the person. To address this, we developed an end-to-end adversarial framework in the digital domain, framing the design of adversarial clothing textures as a 2D texture optimization problem. By crafting an effective adversarial loss function, the framework iteratively generates optimal textures through backpropagation. Our results demonstrate that the textures identified in the digital domain possess strong adversarial properties. Furthermore, we translated these digitally optimized textures into physical clothing and tested them in real-world scenarios, successfully demonstrating that the designed textures significantly degrade the performance of event-based pedestrian detection models. This work highlights the vulnerability of such models to physical adversarial attacks.
\end{abstract}

%
\section{Introduction}
In the 30 milliseconds between frames captured by a traditional camera, a car traveling at 60 kilometers per hour can move approximately 0.5 meters. This significant frame delay makes traditional cameras unsuitable for applications requiring real-time perception and rapid response, such as pedestrian detection in traffic scenarios. In contrast, event cameras operate with an asynchronous triggering mechanism, offering ultra-low latency in the microsecond range and a wide dynamic range ($\geq$ 120 dB). These characteristics make event cameras a superior solution for pedestrian detection tasks, providing lower latency and faster response times compared to traditional cameras.

\begin{figure}[t]
  \centering
  \includegraphics[width=0.9\columnwidth]{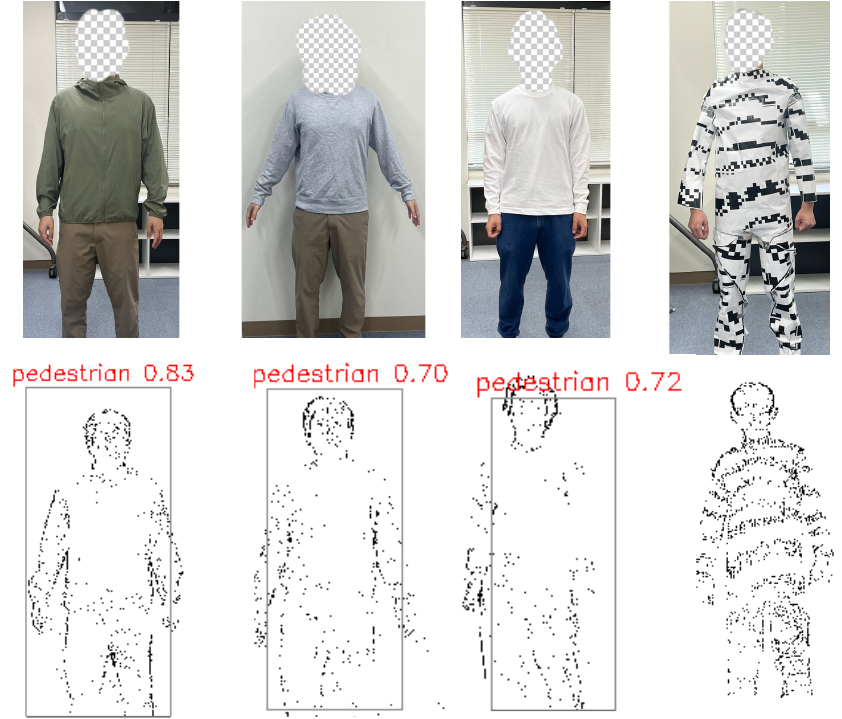}
  \caption{Demonstration of a physical adversarial attack: A person wearing adversarial clothing evades detection by an event-based pedestrian detector during movement, while a pedestrian in normal clothing is accurately detected. Bounding boxes indicate successful pedestrian detection.}
  \label{fig:demostrate}
\end{figure}

In recent years, research on event-based pedestrian detection using deep learning methods has garnered significant attention, particularly with the introduction of large-scale datasets like 1MpX~\cite{perot2020learning} and Gen1~\cite{de2020large}. Notably, RVT~\cite{gehrig2023recurrent} has highlighted the advantages of event cameras in pedestrian detection within traffic scenarios, achieving impressive results by processing event data alone. Further advancements in detection accuracy have been realized through innovative network architectures in models such as HMNet~\cite{hamaguchi2023hierarchical}, GET~\cite{peng2023get}, and SAST~\cite{peng2024scene}. However, the primary focus of current research has been on improving detection performance, with limited attention given to the potential vulnerabilities of event-based pedestrian detectors.

Given that these models are built on deep learning techniques, and considering the well-documented vulnerabilities of deep learning models in the RGB domain~\cite{xu2021towards, cai2022context, carlini2017towards}, it is plausible that event-based pedestrian detectors may exhibit similar security weaknesses. In response, we conducted an unexplored investigation into the security of event-driven pedestrian detection models. 
In this way, event-based vision systems can enhance convenience without compromising human safety.
Unlike previous studies~\cite{marchisio2021dvs, lee2022adversarial}, which primarily focus on digital adversarial attacks by modifying event data, our work is the first to explore the impact of physical adversarial attacks on these detection models. Specifically, we examine how the clothing style of pedestrians can affect the performance of event-driven pedestrian detectors.

Figure~\ref{fig:demostrate} illustrates an example of a physical event attack. Our objective is to design clothing textures that can deceive event-based pedestrian detectors, rendering the wearer undetectable. To better simulate real-world scenarios, we developed an end-to-end digital adversarial attack framework. This framework utilizes 3D differentiable rendering techniques to transform the challenge of designing adversarial textures for physical attacks into a 2D texture optimization problem. By crafting a loss function tailored for adversarial attacks, the framework iteratively generates optimal 2D textures through backpropagation, facilitating the execution of the attack.
To validate the effectiveness of the physical attack, we transferred the optimal texture derived in the digital domain into the physical world and conducted experiments, successfully executing the physical adversarial attack.
Our goal in identifying the adversarial texture is to better develop defenses. This includes designing robust defense algorithms, integrating other types of sensors, and avoiding the fabrication of certain clothing that could interfere with the detection. 

In summary, the contributions of this paper are threefold: 
\begin{enumerate}
\item We present the first exploration of physical adversarial attacks in event-based vision, specifically validating these attacks in  event-based pedestrian detection tasks.
\item We develop an end-to-end digital adversarial attack framework that transforms the design of 3D clothing textures into a 2D texture optimization problem using 3D differentiable rendering techniques, thereby enabling physical adversarial attacks on target detectors.
\item We demonstrate the effectiveness of the proposed method by successfully executing attacks on event-based pedestrian detectors, both in the digital domain and in real-world scenarios, highlighting the vulnerability of these detectors to physical adversarial attacks.
\end{enumerate}

\section{Related Work}
\subsection{Physical Adversarial Attacks}

\noindent 
Deep learning-based vision models have been shown to be vulnerable to adversarial attacks, which typically involve subtle perturbations in the digital domain that can drastically alter model outputs~\cite{xu2021towards, cai2022context, carlini2017towards}. As research in this area has evolved, there has been growing interest in exploring adversarial attacks within the physical world, where the perturbations are applied to objects or environments rather than digital inputs. These physical adversarial attacks are designed to be robust against real-world variations, such as changes in lighting, viewing angles, and distances, enabling them to effectively compromise vision AI models in practical scenarios~\cite{wei2022physical}. 

Common forms of physical attacks include adversarial patches and stickers, which can be placed on objects or worn by people to fool detection systems. These attacks can be categorized into two main types: white-box and black-box. White-box attacks~\cite{hu2021naturalistic, hu2023physically, tan2021legitimate} require detailed knowledge of the target model, including its architecture and parameters, allowing attackers to craft highly effective adversarial perturbations. Black-box attacks~\cite{li2021adversarial, wei2020heuristic}, on the other hand, do not require such knowledge and instead rely on probing the model's responses to various inputs to generate effective perturbations. In this work, we adopt a white-box approach to conduct physical adversarial attacks, leveraging our understanding of the model's inner workings to design precise adversarial patterns.

\subsection{Adversarial Attacks on Event-based Vision}
While most research on adversarial attacks has focused on RGB-based models, the vulnerabilities of visual AI systems that operate in other modalities, such as thermal infrared vision~\cite{zhu2021fooling, zhu2022infrared, wei2023hotcold}, near-infrared vision~\cite{niu2023physics}, and event-based vision~\cite{marchisio2021dvs, lee2022adversarial}, remain relatively underexplored. Event-based vision, in particular, offers unique challenges and opportunities for adversarial attack research due to its asynchronous, high-temporal-resolution nature.

Although research~\cite{hao2023threaten, bu2023rate} on adversarial attacks in spiking neural networks (SNNs) is applicable to event data, this paper focuses specifically on adversarial attacks targeting deep learning models based on event-based data, extending beyond SNNs. For instance, Marchisio et al.~\cite{marchisio2021dvs} explored the creation of adversarial examples by projecting event-based data onto 2D images, generating 2D adversarial images. However, this approach indirectly targets event data and does not directly manipulate the raw input of event cameras, thereby limiting its effectiveness and applicability in real-world scenarios.
Building on this, Lee et al.~\cite{lee2022adversarial} developed an adversarial attack algorithm specifically designed for event-based models. Their approach involved shifting the timing of events and generating additional adversarial events to deceive the model. This method was tested successfully on the N-Caltech101 dataset~\cite{orchard2015converting}, demonstrating the potential for digital adversarial attacks on event-based systems. However, these studies have primarily been confined to the digital domain, where the adversarial perturbations are applied to the event data directly rather than through physical means.
In this paper, we advance the field by extending adversarial attacks on event-based vision into the physical domain, with a focus on pedestrian detection tasks. We design adversarial clothing textures that can deceive event-based pedestrian detectors in real-world scenarios, demonstrating the feasibility and effectiveness of physical adversarial attacks on event-based vision systems.

\begin{figure*}[th]
  \centering
  \includegraphics[width=1\linewidth]{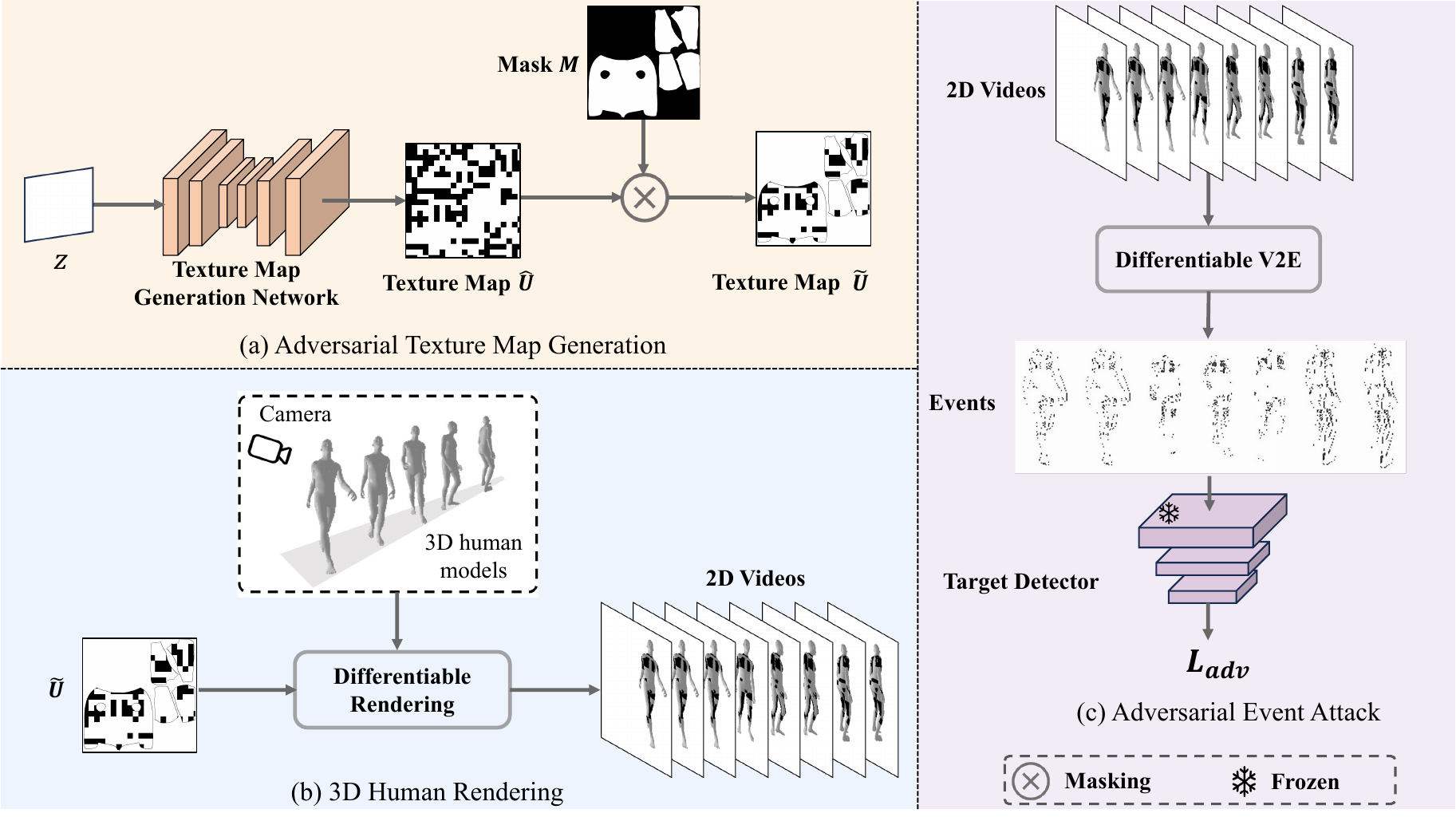}
  \caption{Demonstration of our method. (a) Adversarial Texture Map Generation: The input $z$ is fed into a texture map generation network, producing a grayscale texture map $\hat{U}$. After applying a masking operation, the adversarial texture map $\tilde{U}$ is obtained. (b) 3D Human Rendering: The adversarial texture map $\tilde{U}$ is combined with 3D human model shape and pose parameters, and a differentiable renderer is used to generate 2D videos of continuous human motion. (c) Adversarial Event Attack: From these generated 2D videos, events $\tilde{E}$ are created using the differentiable V2E method and are used to attack event-based pedestrian detectors $f$, where the neural network parameters of $f$ remain frozen. By applying the adversarial loss $L_{adv}$, the entire {end-to-end} pipeline is updated through {backpropagation}, ultimately resulting in the optimal adversarial texture map $\tilde{U}$.
}
  \label{fig:framework}
  \end{figure*}

\section{Methodology}
\subsection{Formulation}
\customparagraph{Event Representation.}
An event stream consists of a series of events, with each event \( e_i \) characterized by the following elements: position \((x_i, y_i)\), timestamp \(t_i\), and polarity \(p_i\). A positive polarity \(p_i = +1\) indicates that the event is triggered when the logarithm of the pixel's intensity increases beyond a positive contrast threshold. Conversely, a negative polarity \(p_i = -1\) indicates that the event is triggered when the logarithm of the pixel's intensity decreases beyond a negative contrast threshold. In this paper, we treat the positive and negative contrast thresholds equally, denoting them collectively as \(\theta\). 
Following the approach used in RVT~\cite{gehrig2023recurrent}, we represent the event stream  ${E}$ over the time interval \([t_1, t_2)\) as follows:
\begin{equation}
E(p, \tau, x, y) = \sum_{e_i \in {E}} \delta(p - p_i) \delta(x - x_i, y - y_i) \delta(\tau - \tau_i), 
\end{equation}
where $\tau_i = \left\lfloor \frac{t_i - t_1}{t_2 - t_1} \cdot B \right\rfloor$ with $B$ as the discretized time bins, and $\delta(x)$ stands for Dirac delta function.

\customparagraph{Problem Definition.}
Let $f$ denotes the pre-trained event-based pedestrian detection model. 
Given the events $E$ as input, the outputs $Y$ of the model consist of the bounding box position $f_{\text{pos}}(E)$, the object probability $f_{\text{obj}}(E)$, and the class score $f_{\text{cls}}(E)$, similar to most object detectors. This relationship is formulated as:
\begin{equation}
    Y = f(E) = [f_{\text{pos}}(E), f_{\text{obj}}(E), f_{\text{cls}}(E)].
\end{equation}
Our goal is to fool the detector so that it fails to detect pedestrians. Specifically, we aim to simultaneously reduce both the object probability and the class score of pedestrians:
\begin{equation}
\min f_{\text{conf}}(E) = \min (f_{\text{obj}}(E) + f_{\text{cls}}(E)).
\label{eq:min}
\end{equation}
To achieve this, we develop a neural network $G$ that generates a 2D texture map $\tilde{U}$ containing adversarial patches. This process involves first creating an initial texture map $\hat{U}$ from a parameter $z$, and then applying a mask $M$ to obtain the expected texture map:
\begin{equation}    
\tilde{U} = {Filter}(G(z), M).
\end{equation}
Utilizing the 3D human model shape $\beta$,  continuous pose parameters set $\boldsymbol{\phi}$, the texture map mask $M$, and the camera extrinsic parameters $[R|t]$, we generate a sequence of 2D frames $\mathbf{I}$ through a differentiable rendering process $\mathcal{R}$. 
These frames represent a human figure wearing clothes with adversarial patches:
\begin{equation}
    \mathbf{I} = \{I_k\}_{k=1}^{N} = \mathcal{R}(\tilde{U}, \beta, \boldsymbol{\phi}, [R|t]),
\end{equation}
where $N$ is the number of frames.
The corresponding adversarial events  $\tilde{E}$ are then  generated using a differentiable video-to-event (V2E) method $T$, based on the event rendering times $\mathbf{t} = \{t_k\}_{k=1}^{N}$:
\begin{equation}
    \tilde{E} = T(\mathbf{I}, \mathbf{t}).
\end{equation}
Therefore, the objective function from Equation~\ref{eq:min} can be reformulated as:
\begin{equation}
\arg\min f_{\text{conf}}(\tilde{E}).
\end{equation}
Here, \(\tilde{E}\) is defined as:
\begin{equation}
    \tilde{E} = T(\mathcal{R}(Filter(G(z), M), \beta, \phi, [R|t]), \mathbf{t}).
\label{eq:objective}
\end{equation}
%
The proposed method aims to identify the optimal adversarial texture map $\tilde{U}$, which generates adversarial events $\tilde{E}$ to deceive the event-based pedestrian detector $f$.

\subsection{Adversarial Attack Framework}
In this section, we provide a detailed overview of our adversarial attack pipeline, as illustrated in Figure~\ref{fig:framework}. 
Our method consists of three key components: adversarial texture map generation, 3D human model rendering, and adversarial event attack. The adversarial texture map generation involves designing the texture pattern and developing the texture map generation network. The adversarial event attack includes the use of a differentiable V2E conversion method and the formulation of an adversarial loss function. By integrating these components, our approach systematically incorporates adversarial patches into event sequences, successfully misleading the event-based pedestrian detector.

\begin{figure}[tb]
  \centering
\includegraphics[width=1\linewidth]{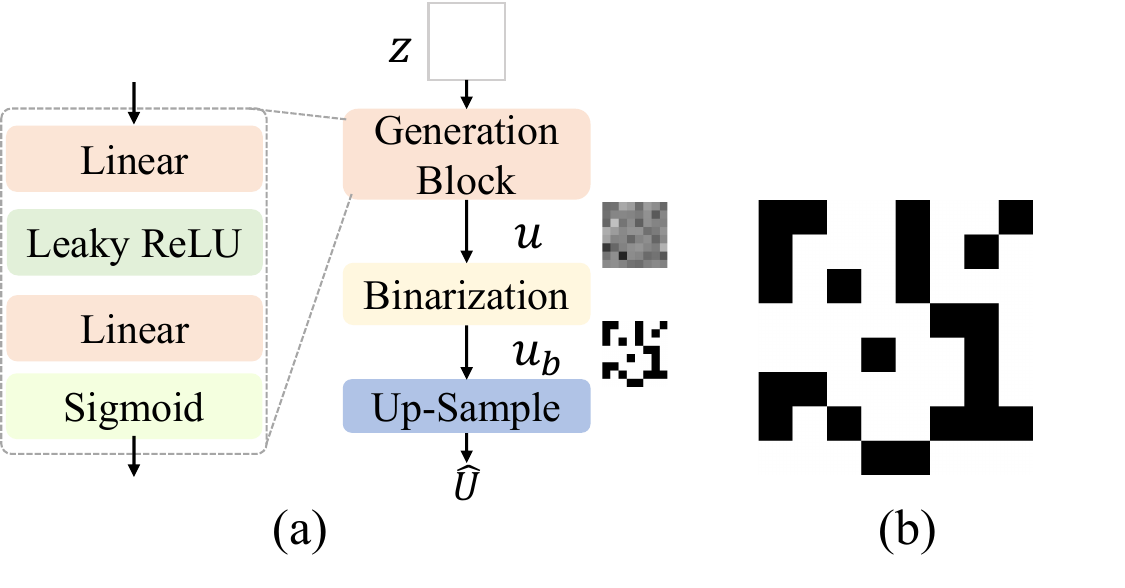}
  \caption{Illustration of the texture map generation network. (a) represents the network structure of the texture map generation network. (b) shows a demo of a texture map $\hat{U}$. The output   $\hat{U}$ consists of an $n \times n$ grid of white or black blocks, where each block is $c\times c$ pixels in size. }
  \label{fig:uvnet}
  \end{figure}

\customparagraph{Design of Texture Map Pattern.}  
Since event cameras capture only changes in brightness, we simplify the design of clothing patterns for the attack by focusing on high-contrast color blocks, specifically black and white, without the need for colored grids. The pedestrian model’s clothing texture is represented using two color blocks: black for low brightness areas and white for high brightness areas. The texture map $\hat{U}$ shown in Figure~\ref{fig:uvnet} (b) illustrates the pattern we designed.

\customparagraph{Texture Map Generation Network.} 
The texture map used in this paper has a resolution of $H \times W$, where $H = W$. It consists of an $n \times n$ grid of blocks, each $c \times c$ pixels, where $c = \frac{H}{n}$.
As shown in Figure~\ref{fig:uvnet}, we develop a texture map generation network that takes $z$ as input, where $z$ is a single-channel $n \times n$ image initialized to white (\ie, value = 1). 
After passing through the generation block, it outputs an $n \times n$ grayscale matrix $u$ with values ranging from 0 to 1. 
This matrix is then binarized, resulting in an $n \times n$ texture map $u_b$ composed solely of black and white (\ie, values are either 0 or 1). 
The final step involves an upsampling operation by a factor of $c$, producing the texture map $\hat{U}$ with a resolution of $H = W$.
%
The binarization operation is implemented using the Straight-Through Estimator (STE)~\cite{bengio2013estimating}, which allows the threshold function to be applied during the forward pass, while using approximate gradients during the backward pass. In the forward pass of STE, we perform the binarization using the following formula:
\begin{equation}
{u_b}_j = 
\begin{cases}
1, & \text{if } {u_j} > 0.5 \\
0, & \text{otherwise} 
\end{cases},
\end{equation}
where $j \in [1, n \times n]$, $u_j$ is the $j$-th value in $u$, and ${u_b}_j$ is the corresponding $j$-th value in ${u_b}$. In practice, to ensure training stability, we avoided direct hard binarization in the STE process. Instead, we adopted a soft binarization approach, gradually transitioning to hard binarization as the training progressed.

To increase rendering flexibility, such as excluding specific regions like the head, feet and fingers from texture rendering, we apply masking to occlude the corresponding areas in the $\hat{U}$, resulting in the texture map $\tilde{U}$, as shown in Figure~\ref{fig:framework}. The masking process is formalized as follows:
\begin{equation}
    \tilde{U}(x, y)  = M(x, y) \cdot \hat{U}(x, y) + (1 - M(x, y)) \cdot \mathbf{1},
\end{equation}
where $\mathbf{1}$ represents a white image of the same size as the mask $M$.

\customparagraph{3D Human Model Rendering.}
One of the key ideas of the proposed attack is to seek for a feasible pattern bounded to a 3D human with the adversarial loss based on 2D detection, in an end-to-end manner.
To this end, we employ differentiable rendering to generate consecutive frames of one SMPL-based 3D human parameterized model~\cite{bogo2016keep}, enabling the back-propagation of gradients from 2D coordinates to the texture map.
The 3D poses at different timestamps are animated from one canonical SMPL model so that the temporal consistency of UV mapping can be guaranteed.
Specifically, we employ the implementation of PyTorch3D~\cite{ravi2020pytorch3d} for differentiable rendering for its native compatibility with PyTorch.
%

\customparagraph{Video to Event.}
To convert a 2D video clip into event data, we begin by transforming a sequence of \( N \) continuous images \(\{I_k\}_{k=1}^{N}\) from the RGB color space to the YUV color space. We then extract the image sequence of the Y channel, which represents the luminance values, and convert these values into logarithmic space.
Given a series of rendering times \(\{t_k\}_{k=1}^{N}\) and a predefined contrast threshold, we calculate the differences between each pair of consecutive frames (\eg, \(I_k\) and \(I_{k+1}\)). 
Subsequently, we compute, in parallel, the number of all positive events \(N^p_{t_{k+1}}\) and negative events \(N^n_{t_{k+1}}\) between these frames.
The differentiable event rendering mechanism we employ is adapted from \cite{gu2021learn}, which makes the process of V2E differentiable by using a near-parallel event rendering reformulation. 
In contrast to the original paper, we fix the contrast threshold at $\theta = 0.2$ rather than using learnable thresholds.

\begin{table*}[t]
\centering
\renewcommand{\arraystretch}{1.1}
\begin{tabular}{c|c|ccc|cccccc}
\toprule
\rowcolor{gray!10}
& \textbf{Confidence} & \multicolumn{3}{c|}{\textbf{Compared Texture}}&\multicolumn{6}{c}{\textbf{Ours Adversarial Texture (\textit{size of grids})}}\\ \cline{3-11} 
\rowcolor{gray!10}
\textbf{Metrics}  & \textbf{Thresholds}                                            & \textbf{White} & \textbf{Black} & \textbf{Random}  & \multicolumn{1}{c}{\begin{tabular}[c]{@{}c@{}}\textbf{60$\times$60}\end{tabular} } & \multicolumn{1}{c}{\begin{tabular}[c]{@{}c@{}}\textbf{50$\times$50}\end{tabular} } &
                                  \multicolumn{1}{c}{\begin{tabular}[c]{@{}c@{}}\textbf{40$\times$40}\end{tabular} }  &\multicolumn{1}{c}{\begin{tabular}[c]{@{}c@{}}\textbf{30$\times$30}\end{tabular} }  &\multicolumn{1}{c}{\begin{tabular}[c]{@{}c@{}}\textbf{20$\times$20}\end{tabular} } 
                                  &\multicolumn{1}{c}{\begin{tabular}[c]{@{}c@{}}\textbf{10$\times$10}\end{tabular} }  \\ \midrule
AP $\downarrow$                               & \multirow{2}{*}{0.001}                          & \multicolumn{1}{c}{  46.8\% }          & \multicolumn{1}{c}{  40.3\%}          & \multicolumn{1}{c|}{ 41.4\% }       & \multicolumn{1}{c}{ 37.2\%}            & \multicolumn{1}{c}{ 39.3\%}            & \multicolumn{1}{c}{ 25.5\%} & \multicolumn{1}{c}{ 17.1\%} &\multicolumn{1}{c}{ \textbf{11.1\%}} &\multicolumn{1}{c}{ 11.8\%}\\ \cline{1-1} \cline{3-11} 
SeqASR $\uparrow$                              &                                            & \multicolumn{1}{c}{0.0\% }          & \multicolumn{1}{c}{0.0\% }          & \multicolumn{1}{c|}{ 0.0\%}       & \multicolumn{1}{c}{0.0\%}            & \multicolumn{1}{c}{0.0\%}            & \multicolumn{1}{c}{0.0\%}& \multicolumn{1}{c}{0.0\%} & \multicolumn{1}{c}{0.1\%} &\multicolumn{1}{c}{ \textbf{0.2\%}} \\ \hline
AP $\downarrow$                               & \multirow{2}{*}{0.01}                          & \multicolumn{1}{c}{  49.7\% }          & \multicolumn{1}{c}{  44.8\%}          & \multicolumn{1}{c|}{ 45.3\% }       & \multicolumn{1}{c}{ 41.0\%}            & \multicolumn{1}{c}{43.4\%}            & \multicolumn{1}{c}{32.6\%} & \multicolumn{1}{c}{ 23.2\%} &\multicolumn{1}{c}{ \textbf{15.7\%}} &\multicolumn{1}{c}{ 15.9\%}\\ \cline{1-1} \cline{3-11} 
SeqASR $\uparrow$                              &                                            & \multicolumn{1}{c}{0.0\% }          & \multicolumn{1}{c}{0.0\% }          & \multicolumn{1}{c|}{ 0.0\%}       & \multicolumn{1}{c}{0.1\%}            & \multicolumn{1}{c}{0.2\%}            & \multicolumn{1}{c}{0.0\%}& \multicolumn{1}{c}{0.4\%} & \multicolumn{1}{c}{\textbf{1.4\%}} &\multicolumn{1}{c}{ \textbf{1.4\%}} \\ \hline
AP $\downarrow$                               & \multirow{2}{*}{0.1}                          & \multicolumn{1}{c}{ 50.6\% }          & \multicolumn{1}{c}{  45.4\%}          & \multicolumn{1}{c|}{ 46.2\% }       & \multicolumn{1}{c}{ 42.2\%}            & \multicolumn{1}{c}{44.5\%}            & \multicolumn{1}{c}{34.4\%} & \multicolumn{1}{c}{22.2\%} &\multicolumn{1}{c}{\textbf{12.2\%}} &\multicolumn{1}{c}{12.5\%}\\ \cline{1-1} \cline{3-11} 
SeqASR $\uparrow$                              &                                            & \multicolumn{1}{c}{0.0\% }          & \multicolumn{1}{c}{0.9\% }          & \multicolumn{1}{c|}{ 0.0\%}       & \multicolumn{1}{c}{1.5\%}            & \multicolumn{1}{c}{1.4\%}            & \multicolumn{1}{c}{9.3\%}& \multicolumn{1}{c}{18.2\%} & \multicolumn{1}{c}{{29.1\%}} &\multicolumn{1}{c}{\textbf{29.8\%}} \\ \hline
AP $\downarrow$   &\multirow{2}{*}{0.25} & \multicolumn{1}{c}{50.5\% }          & \multicolumn{1}{c}{43.8\% }          & \multicolumn{1}{c|}{ 45.2\%}       & \multicolumn{1}{c}{39.4\%}            & \multicolumn{1}{c}{41.7\%}            & \multicolumn{1}{c}{28.0\%}  & \multicolumn{1}{c}{ 14.4\% }  &\multicolumn{1}{c}{ 6.1\%}&\multicolumn{1}{c}{ \textbf{5.7 \%}} \\ \cline{1-1} \cline{3-11} 
SeqASR $\uparrow$                              &                                            & \multicolumn{1}{c}{1.2\%}          & \multicolumn{1}{c}{3.7\%}          & \multicolumn{1}{c|}{4.7\%}       & \multicolumn{1}{c}{8.0\%}            & \multicolumn{1}{c}{9.0\%}            &  \multicolumn{1}{c}{24.7\%}  & \multicolumn{1}{c}{43.2\%}  &\multicolumn{1}{c}{59.5\%}&\multicolumn{1}{c}{ \textbf{63.1\%}} \\ \bottomrule
\end{tabular}
\caption{Result of digital adversarial attack.  The best performance is highlighted in \textbf{bold}.}
\label{tab:result_digital_attack}
\end{table*}

\customparagraph{Adversarial Loss.}
Following Equation~\ref{eq:objective}, we develop the adversarial loss \(\mathcal{L}_{adv}\) to simultaneously minimize both the object confidence score and the class confidence score.
Specifically, the $L_{obj}$ object confidence score is:
\begin{align}
   L_{obj}  =  \frac{1}{M} \sum_{i=1}^{M} f_{obj}^{(i)}(\tilde{E}),
\end{align}
where $M$ is the number of attacked event sequences.
The $L_{cls}$ class confidence score is:
\begin{align}
   L_{cls} =  \frac{1}{M} \sum_{i=1}^{M} f_{cls}^{(i)}(\tilde{E}).
\end{align}
The total loss $L_{adv}$ is the sum of these two losses, 
\begin{align}
L_{adv} = \lambda_1 L_{obj} + \lambda_2 L_{cls},
\end{align}
where $\lambda_1$ = $\lambda_2$ = 10,000 in the experiments.
Using this adversarial loss, we employ back-propagation to iteratively update the adversarial texture maps.

\section{Experiments}
\subsection{Settings}
\customparagraph{Metrics.}
%
We rely on two metrics to evaluate the effectiveness of the adversarial attack. The first is Average Precision (AP), a standard metric in detection tasks, where a lower AP indicates stronger adversarial performance. Since the target detector processes event sequences that record pedestrians, we are particularly interested in its ability to correctly identify pedestrians from these sequences. To capture this, we introduce a sequence-focused metric, the Sequence Attack Success Rate (SeqASR), defined as:
\begin{equation}
\text{SeqASR} = 1 - \frac{N}{M},
\end{equation}
where $M$ is the total number of event sequences, and $N$ is the number of sequences in which pedestrians are successfully detected. A higher SeqASR indicates a greater likelihood that the pedestrian detector fails to correctly identify pedestrians, demonstrating a more effective attack.

\customparagraph{Implementation Details.} 
We use RVT~\cite{gehrig2023recurrent} as the target detector, known for its excellent detection performance with event sequences. Specifically, we select the official pre-trained baseline model RVT-B for our adversarial attack experiments. This detection model was trained on the Gen1 automotive detection dataset~\cite{de2020large}, which consists of real-world event data captured by event cameras at a resolution of 304 $\times$ 240. During both the training and evaluation phases, we keep the target detector frozen.
To ensure compatibility with the target detector, the rendered 2D videos and events are also set to a resolution of 304 $\times$ 240. The texture map used in this study has a resolution of 1024 $\times$ 1024. The time bins $B$ are set to 10. We train the network for 11,500 iterations using the Adam optimizer with an initial learning rate of $10^{-4}$. Our framework is trained and validated with a batch size of 1 on an NVIDIA 3090 GPU.

\customparagraph{Dataset in the Digital Space.}
We utilize the CMU motion capture dataset provided by AMASS~\cite{AMASS:2019}, which includes various subjects in different poses. Each subject comprises several trials, and we select a subset of these trials for our experiments. Our training dataset consists of 47 trials, encompassing 114,173 poses, while the test dataset includes 13 trials, with a total of 10,323 poses.
To enhance the robustness of our attack model, we apply data augmentation by introducing randomness to the camera's extrinsic parameters, varying the angles and sizes of the rendered human figures during training. For comprehensive evaluation, the test set includes 3D human renderings at three different scales.
%

\subsection{Results}
\begin{figure}[tb]
  \centering
\includegraphics[width=1\linewidth]{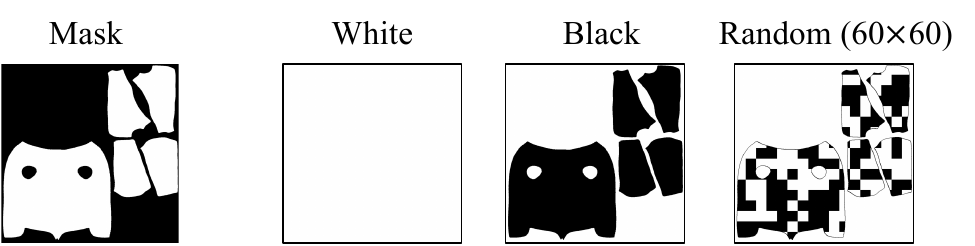}
  \caption{The masks used for all textures and the basic texture maps for comparison.}
  \label{fig:demo_basic_uvmaps_4_compare}
\end{figure}

\customparagraph{Digital Attack.}
Since the evaluation metrics are tied to the confidence thresholds used in the detector's post-processing stage, which can vary depending on the specific detection task, we selected four thresholds between 0.001 and 0.25—common values in detection tasks—to determine the optimal pattern and evaluate attack performance at different confidence levels. As shown in Table~\ref{tab:result_digital_attack}, these thresholds are 0.001, 0.01, 0.1, and 0.25, and the sequence length of the input event is 3. The mask selected for validation is shown in Figure~\ref{fig:demo_basic_uvmaps_4_compare}. To better illustrate performance, we selected three texture maps for comparison: white, black, and a random pattern (60 $\times$ 60 pixels), as also illustrated in Figure~\ref{fig:demo_basic_uvmaps_4_compare}.

\begin{figure}[tb]
  \centering
\includegraphics[width=0.9\linewidth]{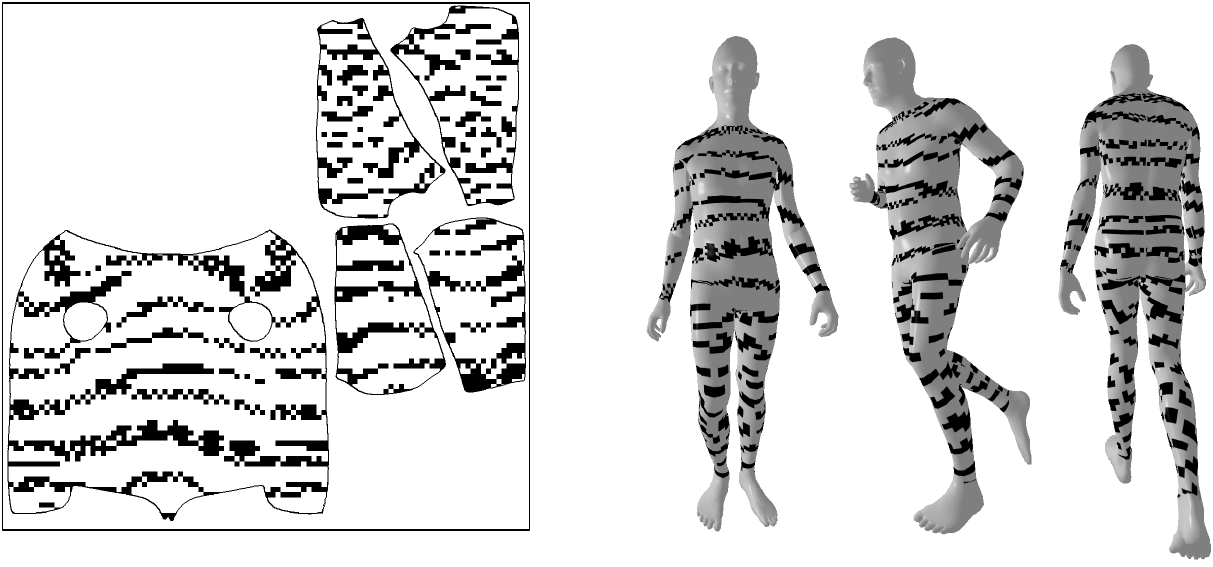}
  \caption{Visualization of the optimal texture map (grid size 10$\times$10 pixels) and the corresponding rendered human.}
  \label{fig:best_pattern_10x10}
  \end{figure}

\begin{figure}[t]
  \centering
  \includegraphics[width=1\linewidth]{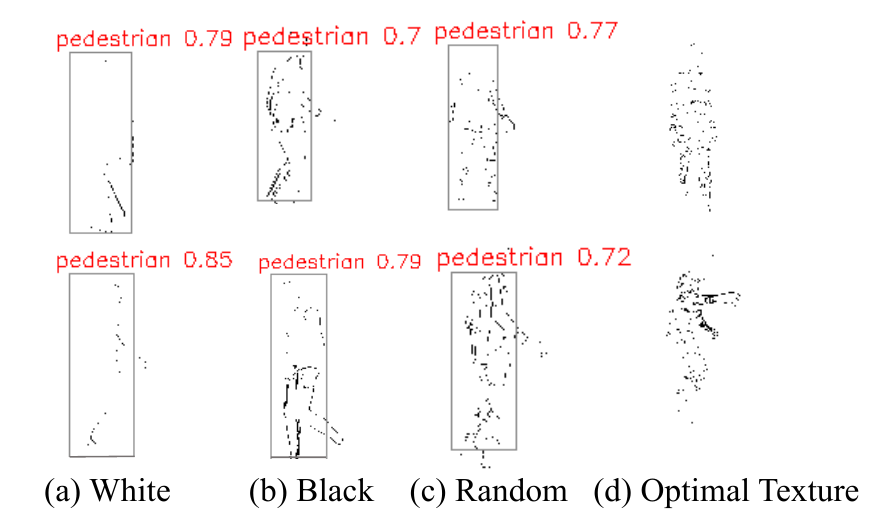}
  \caption{Visualization of digital attacks. Bounding boxes indicate the successful detection of pedestrians.}
  \label{fig:vis_physical_attack_event}
\end{figure}

%
We trained six different grid sizes, ranging from 60$\times$60 to 10$\times$10. The test results, shown in Table~\ref{tab:result_digital_attack}, indicated that the 20$\times$20 and 10$\times$10 grids achieved better AP and SeqASR scores. We selected the 10$\times$10 grid, which performed better in the SeqASR metric, as the optimal texture pattern for subsequent experiments. The optimal 10$\times$10 texture pattern is shown in Figure~\ref{fig:best_pattern_10x10}. The visualization of prediction results for different texture maps is shown in Figure~\ref{fig:vis_physical_attack_event}, demonstrating that the optimal texture can successfully deceive the event-based pedestrian detector in the digital domain.

To further investigate the impact of textures on different body parts (including the upper body, legs, and arms) on overall attack performance, we selected the 10$\times$10 grid size and trained with various mask combinations to identify the optimal pattern for each combination. The confidence threshold was set at 0.25 during training and testing. The final evaluation results are presented in Table~\ref{tab:result_digital_attack_bodypart}. The quantitative results indicate that the lower body (i.e., legs) exhibits the best attack performance among the three regions. Performance generally improves as more body parts are covered by the adversarial texture, with full-body coverage yielding the most significant impact, which aligns with our expectations.

\begin{table}[t]
\centering
\renewcommand{\arraystretch}{1.1}
\begin{tabular}{ccc|cc}
\toprule
\rowcolor{gray!10}
\multicolumn{3}{c|}{{\textbf{Body Parts in Texture Map}}} & \multicolumn{2}{c}{\textbf{Metrics}}\\ \hline
\rowcolor{gray!10}
\multicolumn{1}{c}{upper body} & \multicolumn{1}{c}{arms} & \multicolumn{1}{c|}{legs} & AP $\downarrow$ & SeqASR $\uparrow$ \\ \midrule
 \multicolumn{1}{c|}{ \checkmark}  & \multicolumn{1}{c|}{ }     & \multicolumn{1}{c|}{ }       & 44.6\% &4.7\%\\  \cline{1-5}  
  \multicolumn{1}{c|}{ }  & \multicolumn{1}{c|}{\checkmark }     & \multicolumn{1}{c|}{ }       & 49.7\% &1.2\%\\  \cline{1-5}  
   \multicolumn{1}{c|}{ }  & \multicolumn{1}{c|}{}     & \multicolumn{1}{c|}{\checkmark }       & 30.1\% &21.9\%\\  \cline{1-5}  
    \multicolumn{1}{c|}{\checkmark}  & \multicolumn{1}{c|}{\checkmark }     & \multicolumn{1}{c|}{ }       & 40.8\% &5.6\%\\  \cline{1-5}  
    \multicolumn{1}{c|}{\checkmark}  & \multicolumn{1}{c|}{ }     & \multicolumn{1}{c|}{\checkmark}       & 11.9\% &52.0\%\\  \cline{1-5}  
    \multicolumn{1}{c|}{}  & \multicolumn{1}{c|}{\checkmark }     & \multicolumn{1}{c|}{\checkmark}       & 21.6\% &30.3\%\\  \cline{1-5} 
  \multicolumn{1}{c|}{\checkmark }  & \multicolumn{1}{c|}{\checkmark }     & \multicolumn{1}{c|}{\checkmark }       & \textbf{5.7\%} & \textbf{63.1\%} \\ \bottomrule
\end{tabular}
\caption{Result of digital adversarial attack with different rendered body parts. 
}
\label{tab:result_digital_attack_bodypart}
\end{table}

\begin{figure}[t]
  \centering
  \includegraphics[width=0.8\linewidth]{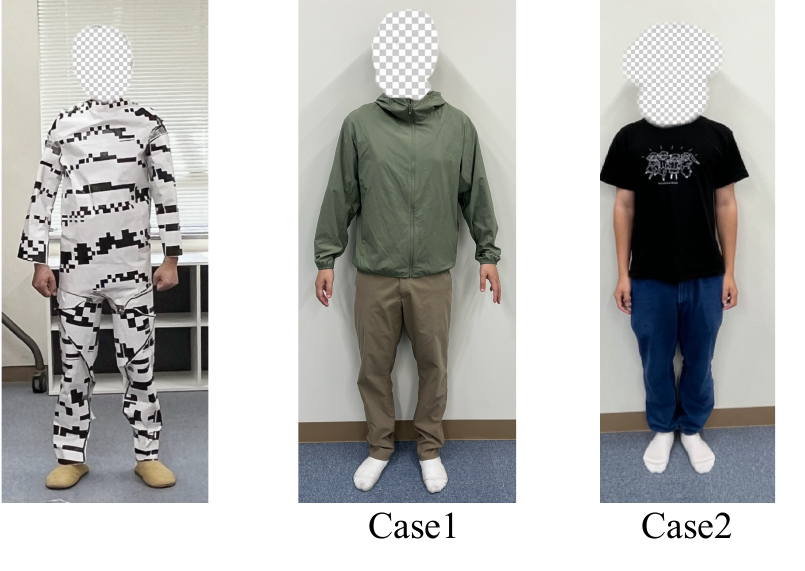}
  \caption{Visualization of the clothes with optimal texture and the compared clothes for validation.}
  \label{fig:demo_clothes}
  \end{figure}

\begin{figure}[t]
  \centering
  \includegraphics[width=1\linewidth]{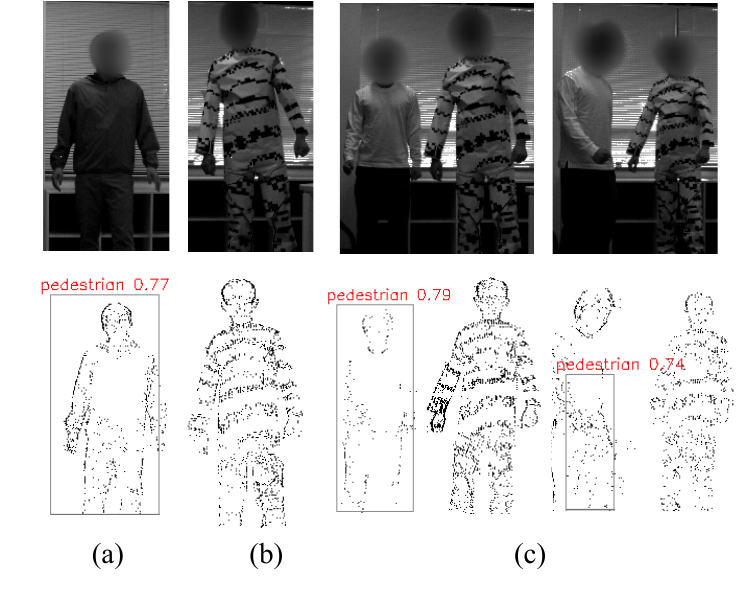}
  \caption{Visualization of physical attacks in indoor scenes. Bounding boxes indicate the successful detection of pedestrians. Images are cropped for better visualization.}
  \label{fig:vis_physical_attack}
\end{figure}  

\begin{table}[t]
\centering
\renewcommand{\arraystretch}{1.2}
\resizebox{0.97\linewidth}{!}
{
\begin{tabular}{c|c|cc|c|c}
\toprule
\rowcolor{gray!10}
 & \textbf{Confidence}   & \multicolumn{3}{c|}{\textbf{Indoor}} & \textbf{Outdoor}\\ \cline{3-6} 
 \rowcolor{gray!10}
 \textbf{Metrics}&         \textbf{Thresholds}       & \multicolumn{1}{c}{\textbf{Case1}} & \multicolumn{1}{c}{\textbf{Case2}}  & \multicolumn{1}{c|}{\textbf{Ours}} &\multicolumn{1}{c}{\textbf{Ours}} \\ \midrule         
AP $\downarrow$ &\multirow{2}{*}{0.001} & 9.1\% & 15.1\% & \textbf{0.0\%} & {7.6\%}\\ \cline{1-1}  \cline{3-6}  
SeqASR $\uparrow$ &  & 0.0\%& 0.0\% &\textbf{0.0\%} & {0.0\%} \\  \cline{1-6}                    
AP $\downarrow$ &\multirow{2}{*}{0.01} & 12.6\% & 16.7\% &   \textbf{0.0\%} & {7.8\%}\\ \cline{1-1}  \cline{3-6} 
SeqASR $\uparrow$ &  & 0.0\%&0\%   & \textbf{2.1\%} & 1.4\% \\  \cline{1-6} 
AP $\downarrow$ &\multirow{2}{*}{0.1} & 15.1\% & 19.2\% & \textbf{0.0\%} & 6.0\% \\ \cline{1-1}  \cline{3-6} 
SeqASR $\uparrow$ &  & 0.0\%& 0.0\%& \textbf{37.9\%} & 21.6\%\\ \cline{1-6} 
AP $\downarrow$ &\multirow{2}{*}{0.25} & 14.5\% & 18.0\%&  \textbf{0.0\%}&{4.6\%}\\ \cline{1-1}  \cline{3-6} 
SeqASR $\uparrow$ &  & 4.3\%& 0.0\% &  \textbf{74.3\%} & {46.0\%} \\ \bottomrule
\end{tabular}}
\caption{Result of physical adversarial attack. }
\label{tab:result_physical_attack}
\end{table}


\begin{figure}[t]
  \centering
\includegraphics[width=1\linewidth]{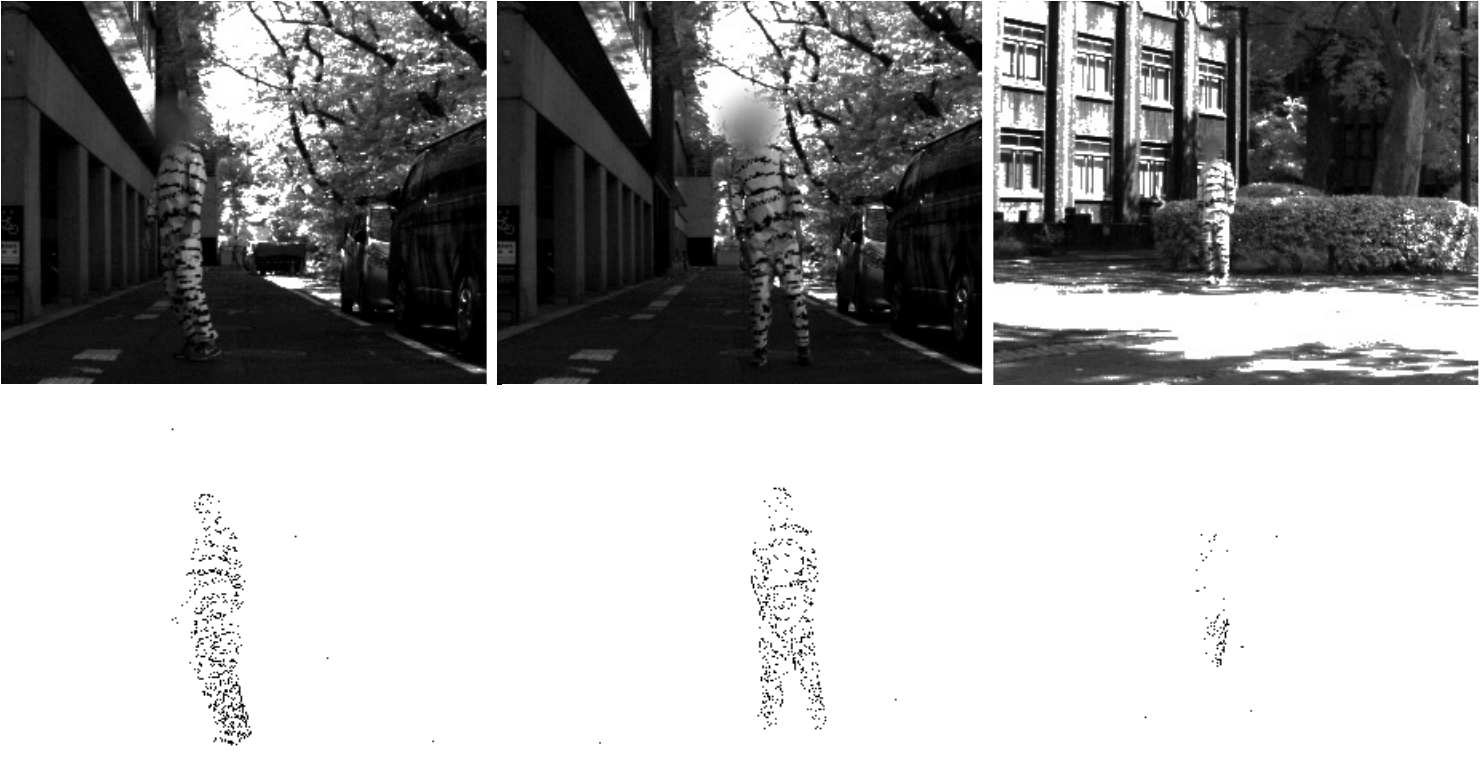}
  \caption{Visualization of the detection results in outdoor scenes. 
  None of the pedestrians is detected.
  }
  \label{fig:outdoor_vis}
\end{figure}

\customparagraph{Physical Attack.}
For the physical attack experiments, we used an INIVATION DAVIS346 MONO event camera to capture real-world event data, with a spatial resolution of 346$\times$260 pixels. The output was cropped to 304$\times$240 pixels to ensure compatibility with the target detector. The input event sequence length for the detector was set to 10. We then expanded the optimal texture pattern obtained from the digital attack experiments and printed it on paper. The printed textures were cut and assembled into clothing pieces according to body parts, as shown in Figure~\ref{fig:demo_clothes}. For comparison, we selected two additional sets of clothing to serve as case studies, also illustrated in Figure~\ref{fig:demo_clothes}.
The event camera was fixed in place to record subjects wearing different textured clothing, with each subject performing the same set of actions. Since the event camera captures grayscale images alongside event data, we used YOLOv7~\cite{wang2023yolov7} to annotate bounding boxes and classify these grayscale images, providing ground truth labels. For each texture, we collected 1,400 consecutive images and evaluated the AP and SeqASR metrics, as detailed in Table~\ref{tab:result_physical_attack}. In indoor scenes, we observed that the ordinary clothing in Case 1 and Case 2 exhibited lower SeqASR, while the optimal texture demonstrated superior adversarial performance in the physical attack. This indicates that the pattern remains effective when transitioning from the digital domain to physical attacks. Figure~\ref{fig:vis_physical_attack} illustrates the impact of clothing textures on event-based detection. In Figure~\ref{fig:vis_physical_attack} (a), the pedestrian wearing normal textures is successfully detected by the event-based detector. Conversely, in Figure~\ref{fig:vis_physical_attack} (b), the adversarial texture effectively evades detection.
Panel (c) highlights the difference between normal and adversarial textures: the former is detected by the detector, while the latter bypasses it entirely.

When considering the optimal texture across both indoor and outdoor scenes, the AP and SeqASR metrics reveal a decrease in the effectiveness of our adversarial clothing in outdoor environments. This reduction can be attributed to the more complex, dynamic backgrounds and fluctuating lighting conditions present in outdoor settings. Despite these challenges, the physical adversarial attack success rate remains notably high even in outdoor scenarios.
Figure~\ref{fig:outdoor_vis} visualizes the detection results in outdoor scenes, showing that a pedestrian wearing adversarial textured clothing effectively evades detection by the event-based pedestrian detector.

\section{Conclusion}
This paper presents an end-to-end method for creating adversarial clothing designed to attack event-based pedestrian detectors in the physical domain. We address the challenge of identifying the most effective adversarial clothing by formulating it as a task of optimizing a 2D texture map in the digital domain. Through the use of 3D rendering techniques, this optimized texture pattern is mapped onto a 3D human model, where it demonstrates strong adversarial effectiveness in the digital space. We then successfully translate this texture pattern into the physical domain, achieving comparable attack results. Our findings indicate that event-based pedestrian detectors, much like their RGB-based counterparts, are vulnerable to security breaches.

\customparagraph{Limitations and Future Work.}
This study focuses on designing black-and-white adversarial patches for texture mapping, but real-world clothing typically features a wider range of colors and more complex textures. While effective, our approach will be extended to incorporate more realistic and diverse clothing styles, aligning better with real-world scenarios. Additionally, we will develop advanced defense mechanisms to counter various physical adversarial attacks.

\section{Acknowledgments}
We express our gratitude to Mingze Ma and Yifan Zhan for their contributions to conducting the experiments.
This research was supported in part by JSPS KAKENHI Grant Numbers 24K22318, 22H00529, 20H05951, JST-Mirai Program JPMJMI23G1, and ROIS NII Open Collaborative Research 2023-23S1201, JST SPRING (Grant Number JPMJSP2108), Guangdong Natural Science Funds for Distinguished Young Scholars (Grant 2023B1515020097), the AI Singapore Programme under the National Research Foundation Singapore (Grant AISG3-GV-2023-011), and the Lee Kong Chian Fellowships. 
\bibliography{aaai25}

\clearpage

\section{More Implementation Details}
\customparagraph{Details of the Differentiable V2E.}
In this paper, we follow the V2E method proposed by \citet{gu2021learn}. Specifically, for a given position \(\mathbf{p}\), we calculate the number of events from the following formula:
\begin{equation}
    \begin{alignedat}{2}
        N^p_{t_{k+1}}(\mathbf{p}) &= h(D_{t_{k+1}}(\mathbf{p})), \\
        N^n_{t_{k+1}}(\mathbf{p}) &= h(-D_{t_{k+1}}(\mathbf{p})), \\
    \end{alignedat}
\end{equation}
where 
\begin{align}
    D_{t_{k+1}}(\mathbf{p}) = S_{t_{k+1}}(\mathbf{p}) - S_{t_k}(\mathbf{p}),
\end{align}
and $h(x)=\max(x,0)$ is the half-rectification function. \(S_{t_{k+1}}(\mathbf{p})\) indicates the accumulated change of polarities:
\begin{equation}
S_{t_{k+1}}(\mathbf{p}) = \left\lfloor \hat{S}_{t_{k+1}}(\mathbf{p}) \right\rfloor + r_{t_{k+1}}(\mathbf{p}),
\end{equation}
where \(r_{t_{k+1}}(\mathbf{p})\) is an adaptive residual term to compensate for rounding errors.
It is important to note that, unlike the original paper, we do not use learnable contrast thresholds but instead fix the contrast threshold as follows:
\begin{equation}
\begin{alignedat}{2}
    \hat{S}_{t_{k+1}}(\mathbf{p}) = \frac{\log(I_{t_{k+1}}(\mathbf{p})) - \log({I}_{t_1} (\mathbf{p}))}{\theta}, \\
\end{alignedat}
\end{equation}
where $\theta$ = 0.2 in this case

\begin{figure}[th]
  \centering
  \includegraphics[width=1\linewidth]{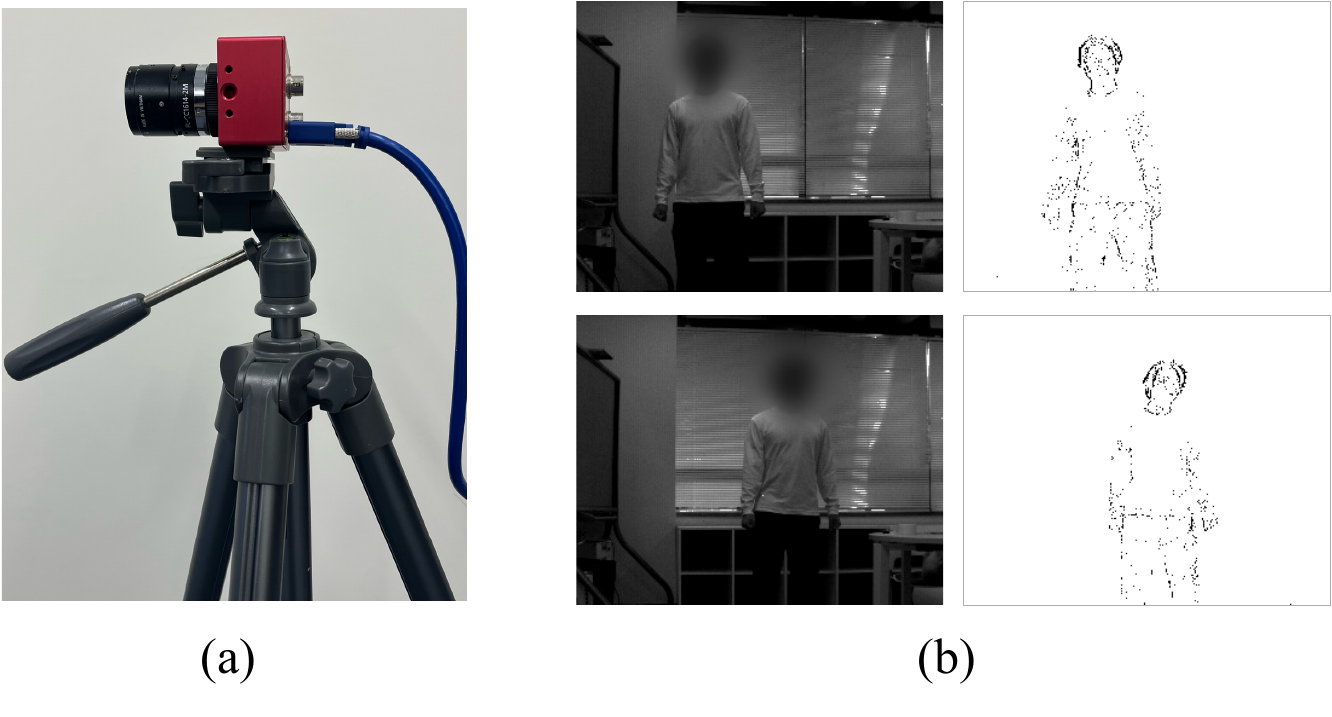}
  \vspace{-20pt}
  \caption{The imaging acquisition system and the visualization of the real-captured data.}
  \label{fig:supp_acquisition_system}
\end{figure}

\customparagraph{Details of the Physical Adversarial Attack Experiments.}
Figure~\ref{fig:supp_acquisition_system} (a) showcases our imaging acquisition system, which is centered around an INIVATION DAVIS346 event camera, outfitted with a 12 $mm$ lens and control signal interfaces. 
This setup captures real-world event data while concurrently acquiring grayscale images at 25 $fps$ (frames per second). 
Both the grayscale images and event data are spatially aligned, each with a resolution of $346\times260$ pixels. We maintained temporal synchronization between the event stream and grayscale images throughout the processing. The original resolution was cropped to $304\times240$ pixels to ensure compatibility with the target detector. Figure~\ref{fig:supp_acquisition_system} (b) shows the cropped event frames and corresponding grayscale images.

\begin{figure}[th]
  \centering
\includegraphics[width=1\linewidth]{ 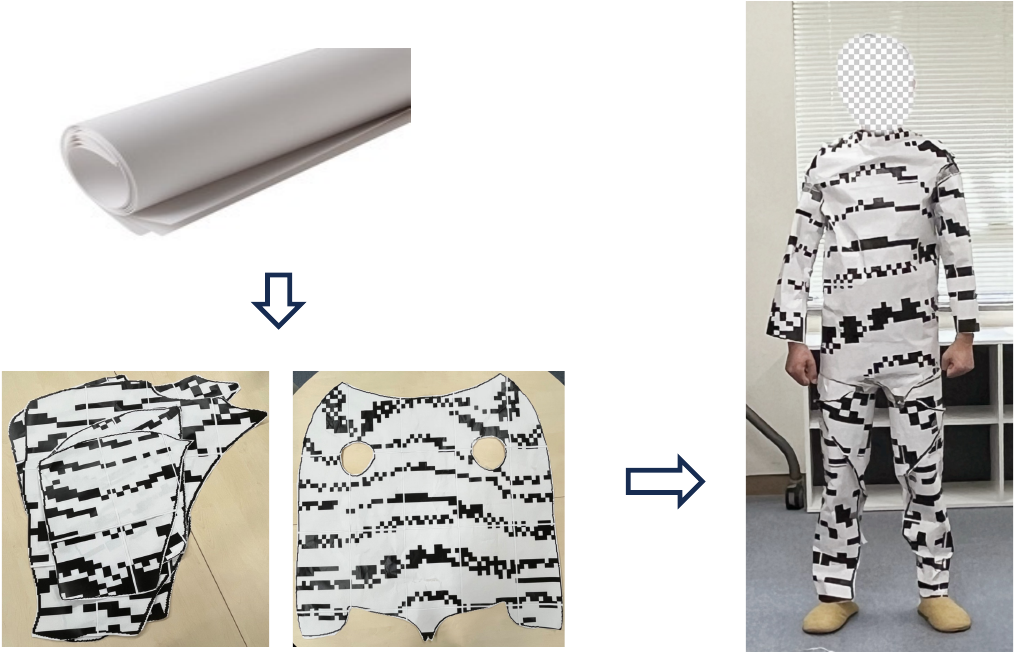}
  \caption{The process of fabricating adversarial textured clothing.}
  \label{fig:supp_clothes_making}
\end{figure}

In the physical adversarial attack experiments, to verify whether the clothing with the optimal adversarial texture obtained in the digital domain could produce real physical adversarial attack effects in the real world, we fabricated the clothing with the optimal adversarial texture. The entire fabrication process is illustrated in Figure~\ref{fig:supp_clothes_making}. We first enlarged the texture pattern and printed it onto white paper, then cut and assembled the pieces according to different body parts. The result was a complete set of clothing with the optimal adversarial texture, which can be naturally worn for physical adversarial attack experiments.

\end{document}